\def\year{2021}\relax
\documentclass[letterpaper]{article} 
\usepackage{aaai21}  
\usepackage{times}  
\usepackage{helvet} 
\usepackage{courier}  
\usepackage[hyphens]{url}  
\usepackage{graphicx} 
\usepackage{natbib}  
\usepackage{caption} 
\usepackage[ruled,linesnumbered]{algorithm2e}
\usepackage[switch]{lineno} 
\usepackage{bm}
\usepackage{amsmath}
\usepackage{multirow}
\usepackage{amssymb}
\usepackage{subfigure}
\usepackage{booktabs}       
\graphicspath{{PIC/}}

\newcommand{\CF}{\mathcal{F}}
\newcommand{\TCF}{\tilde{\mathcal{F}}}

\newcommand{\Bx}{\mathbf{x}}

\newcommand{\BX}{\mathbf{X}}

\newcommand{\BB}{\mathbf{B}}
\newcommand{\BD}{\mathbf{D}}

\newcommand{\BH}{\mathbf{H}}

\newcommand{\BSG}{\bm{\Sigma}}
\newcommand{\BI}{\mathbf{I}}

\newcommand{\BC}{\mathbf{C}}

\newcommand{\Br}{\mathbf{r}}

\newcommand{\CS}{\mathcal{S}}

\newcommand{\CB}{\mathcal{B}}

\newcommand{\CG}{\mathcal{G}}

\newcommand{\CL}{\mathcal{L}}

\newcommand{\bmu}{\bm{\mu}}
\newcommand{\TBx}{\tilde{\Bx}}

\newcommand{\hBx}{\hat \Bx}

\newcommand{\mrRC}{\multirow{4}{*}{\tabincell{c}{ResNet\\ \& \\CIFAR10}}}
\newcommand{\mrVC}{\multirow{4}{*}{\tabincell{c}{VGG\\ \& \\ CIFAR10}}}
\newcommand{\mrRS}{\multirow{4}{*}{\tabincell{c}{ResNet\\ \& \\ SVHN}}}
\newcommand{\mrVS}{\multirow{4}{*}{\tabincell{c}{VGG\\ \& \\ SVHN}}}

\newtheorem{theorem}{$\mathbf{Theorem}$}

\newcommand{\eg}{\textit{e}.\textit{g}.}
\newcommand{\tabincell}[2]{
	\begin{tabular}{@{}#1@{}}#2\end{tabular}
}

\title{Detecting Adversarial Examples from Sensitivity Inconsistency of Spatial-Transform Domain}
\author{
	Jinyu Tian,\textsuperscript{\rm 1}
	Jiantao Zhou,\textsuperscript{\rm 1,}\footnote{The corresponding author.}
	Yuanman Li,\textsuperscript{\rm 2}
	Jia Duan\textsuperscript{\rm 1}
	\\
}
\affiliations{
	\textsuperscript{\rm 1}State Key Laboratory of Internet of Things for Smart City, \\ Department of Computer and Information Science, University of
	Macau\\
	\textsuperscript{\rm 2}Guangdong Key Laboratory of Intelligent Information Processing, \\ College of Electronics and Information Engineering, Shenzhen University
	
	\{yb77405, jtzhou\}@um.edu.mo, yuanmanli@szu.edu.cn, xuelandj@gmail.com
	
}

\begin{document}
\maketitle

\begin{abstract}
	Deep neural networks (DNNs) have been shown to be vulnerable against adversarial examples (AEs), which are maliciously designed to cause dramatic model output errors. In this work, we reveal that normal examples (NEs) are insensitive to the fluctuations occurring at the highly-curved region of the decision boundary, while AEs typically designed over one single domain (mostly spatial domain) exhibit exorbitant sensitivity on such fluctuations. This phenomenon motivates us to design another classifier (called dual classifier) with transformed decision boundary, which can be collaboratively used with the original classifier (called primal classifier) to detect AEs, by virtue of the sensitivity inconsistency. When comparing with the state-of-the-art algorithms based on Local Intrinsic Dimensionality (LID), Mahalanobis Distance (MD), and Feature Squeezing (FS), our proposed Sensitivity Inconsistency Detector (SID) achieves improved AE  detection performance and superior generalization capabilities, especially in the challenging cases where the adversarial perturbation levels are small. Intensive experimental results on ResNet and VGG validate the superiority of the proposed SID.  
\end{abstract}

\begin{figure*}[!t]
	\centering
	\includegraphics[width = 0.6\textwidth]{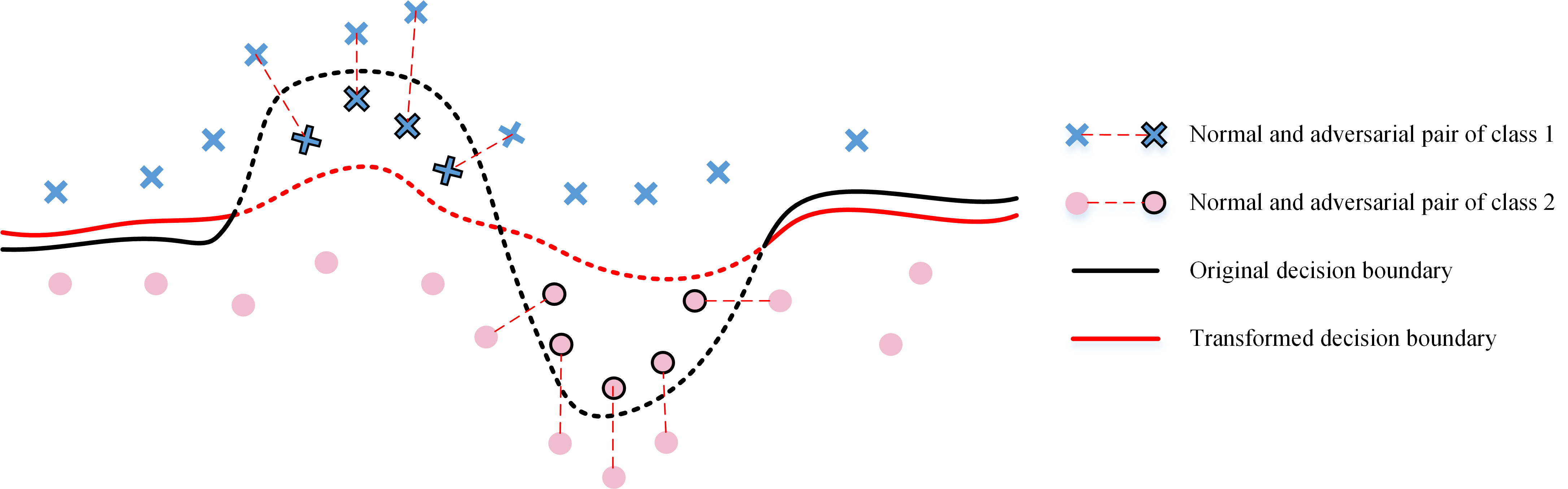}	
	\caption{An illustrative example of the sensitivity of AEs against decision boundary fluctuation. }
	\label{fig:toy}
\end{figure*}

\section{Introduction}
Deep neural networks (DNNs) have achieved the state-of-the-art performance on a wide range of tasks including image classification  \cite{DLApp1}, speech recognition \cite{DLApp2}, etc. However, recent studies have shown that DNNs are vulnerable to crafted adversarial examples (AEs), which are generally imperceptible to the sense of humanity while being able to cause severe model output errors. Such vulnerability leads to serious security risks when deploying DNNs on critical scenarios, \eg, self-driving cars. The attempts for defending against the threat of AEs can be roughly categorized into two types: robust classification and AE detection. The former type aims to eliminate the impact of the adversarial noise and make correct classification (\eg, adversarial training \cite{robustlearning}, denoising, \cite{HGD,DUP}, defensive distillation \cite{Distillation1}, etc.). Though many robust classification strategies have been proposed, most of them are still not powerful enough to defeat the secondary attack or AEs generated by some advanced attacks \eg, C\&W \cite{CW}. In fact, in \cite{LimitRobus}, it was pointed out that the true robustness would lead to depreciation in accuracy. Alternatively, a weaker version of the defense is just to detect the AEs, while not ambiguously rectifying the classification results. In many practical applications, such detection is still quite meaningful, generating alarming signals to potential threats. Further, the AE detection could guide a better defense strategy \cite{GuideDef}.

An AE can be regarded as the translation of a normal example (NE) along the adversarial direction. Geometrically, the adversarial direction usually points toward the highly-curved region of the decision boundary, such that the decision boundary can be crossed with the minimized perturbation magnitude \cite{ClassificationRegion}.  A symmetric phenomenon we would like to point out is that if we \textbf{can intentionally cause fluctuations at the highly-curved regions of the decision boundary}, then those AEs could easily lead to different classification results, while NEs would exhibit quite consistent behavior. Such phenomenon reveals that, for NEs and AEs, there exists an inconsistency of sensitivity to boundary fluctuation at highly-curved regions. This motivates us to design another classifier with transformed decision boundary, which can be collaboratively used with the original classifier (called \textbf{primal classifier}) to detect AEs, by virtue of the sensitivity inconsistency. Ideally, the designed classifier (called \textbf{dual classifier}) should have dissimilar structures at the highly-curved regions with the primal classifier, while maintaining similar structures at the other regions.

To design the dual classifier satisfying the above requirements, we resort to the transform domain techniques, i.e., design the classifier over the transform domain, rather than the traditional spatial domain. In fact, some existing works \cite{Liu2016, Tramer2017} pointed out that models trained on the same feature domain of a given dataset tend to produce similar decision boundaries, especially their curved regions. This would explain why AE has the ability of transferable attack on models with different architectures. More specifically, we construct the dual classifier by using the \emph{Weighted Average Wavelet Transform} (WAWT) and then propose a simple yet effective method for detecting AEs based on the sensitivity inconsistency of NEs and AEs. Our major contributions can be summarized as follows: 1) We reveal the existence of sensitivity inconsistency between NEs and AEs against the decision boundary transformation which is fulfilled by constructing the dual classifier. We theoretically prove the effectiveness of the designed dual classifier in affine and quadratic cases, and empirically demonstrate it for general cases; 2) Motivated by the sensitivity inconsistency, we define a feature to evaluate the sensitivity of an unknown example to the boundary transformation, and then propose a method called \emph{Sensitivity Inconsistency Detector} (SID) to effectively detect AEs; and 3) When comparing with the state-of-the-art algorithms based on Local Intrinsic Dimensionality (LID) \cite{LID}, Mahalanobis Distance (MD) \cite{OOD}, and Feature Squeezing (FS) \cite{FeatureSqueezing}, we observe improved detection performance and superior generalization capabilities, especially in the challenging cases where the perturbation levels are small. Experimental results on ResNet and VGG validate the superiority of the SID. 

\emph{Notations:} Without loss of generality, we restrict our attention to a $K$-class DNN classifier  {\small{$\CF$}} trained on dataset  {\small{$\BX = \{\Bx_i\}_{i=1}^N$}}. For a NE  {\small{$\Bx$, $\CF$}} calculates $K$ logit values, i.e.,   {\small{$\CF (\Bx) =\{f_1(\Bx),f_2(\Bx),...,f_K(\Bx)\}$}}, where  {\small{$f_i(\Bx)$}} is the prediction confidence of  {\small{$\Bx$}} to the $i$-th class. The predicted label of $\Bx$ is denoted by  {\small{$k(\Bx) = \mathtt{argmax}_i \{f_i(\Bx)\}, \mathrm{for}~ i=1,2, \cdots, K$}}. The adversarial counterpart of $\Bx_i$ is written as $\hat \Bx_i = \Bx_i + \Br_i$, where $\Br_i$ is the adversarial perturbation. The set of all adversarial samples is expressed as  {\small{$\hat \BX = \{\hBx_i\}_{i=1}^N$}}. 

\begin{figure*}[!t]
	\centering
	\includegraphics[width = 0.6\textwidth]{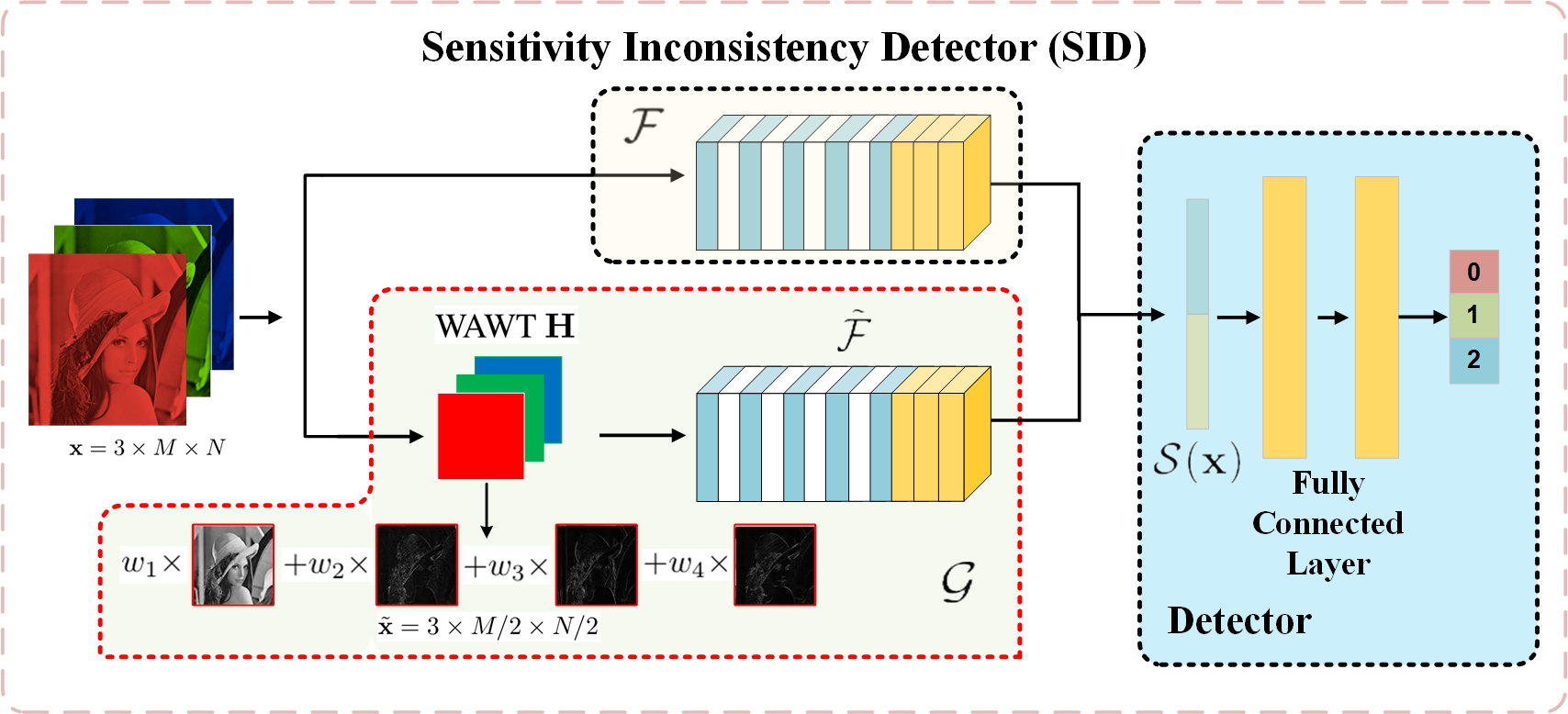}	
	\caption{The schematic diagram of the proposed SID.}
	\label{fig:framework}
\end{figure*}

\section{Related Works}

In this section, we briefly review the state-of-the-art methods on adversarial attack and AE detection. As there are a large number of methods on both topics, we only mention some of the representatives.

\textbf{Adversarial Attack}: The adversarial attack is to craftily manipulate the normal input with imperceptible distortions to make the pre-trained model misclassify. Formally, an AE $\hat \Bx$ targeted on model $\CF$ is essentially a correctly classified example $\Bx$ added with the adversarial perturbation $\Br$ such that $k(\Bx) \neq k(\hat \Bx)$. The added $\Br$ is typically constrained by $L_{p}$ norm \cite{suitabilityLp}, i.e.,

\begin{equation}\label{optimal}
\min_{\Br} \|\Br\|_p ~~~ \text{s.t.} ~~~ k(\Bx+\Br) \neq k(\Bx).
\end{equation}

Several adversarial attacks attempted to approximate the solution of this non-convex optimization problem with different searching and relaxation strategies. Goodfellow \emph{et. al} proposed a method called Fast Gradient Sign Method (FGSM) to search for a feasible solution along the negative gradient sign direction of the cost function with an empirical step \cite{FGSM}. To improve the attack performance, Kurakin \emph{et. al} designed the Basic Iterative Method (BIM) \cite{BIM} by adopting an iterative searching strategy. Further, Moosavi \emph{et al.} suggested a powerful attack DeepFool \cite{DeepFool}, approximating the optimal solution of (\ref{optimal}) based on Newton-Raphson method. In addition, Carlini and Wagner (C\&W) \cite{CW} relaxed the constraint in (\ref{optimal}) with a margin loss, which can be integrated into the objective function and adaptively produce AEs with minimal perturbation levels. Instead of directly solving the problem (\ref{optimal}), Jacobian Saliency Map Attack (JSMA) \cite{JSMA} constructs AEs through a greedy iterative procedure, altering only the pixels which contribute most to the correct classification. Some other recent works on adversarial attacks can be found in \cite{BoundaryAttack, RoadSign, AdvPatch} and references therein.

\textbf{Detection of AEs}: Most of the AE detection methods are based on the observation that AEs lie far from the distribution of NEs \cite{Steganalysis, MeasuresUncertainty, FarAwayDis1,FarAwayDis2, FarAwayDis3, Hidden1, Hidden4,Hidden5}.  Along this line, Jan \emph{et. al} \cite{TDetector1} augmented classification networks by subnetworks branching off the main network at some layers and produced a probability of the input being adversarial. Kathrin \emph{et. al} \cite{STest1} detected AEs by using the statistical test to evaluate the confidence level of the ``belonging'' of an unknown example to NE distribution. Feinman \emph{et al.} \cite{KDBU} proposed two features: Kernel Density (KD) and Bayesian-Uncertainty (BU) to evaluate the proximity of an example to the NE manifold. Ma \emph{et al.} pointed out that KD and BU have limitations of characterizing local adversarial regions. As a remedy, they proposed the feature LID \cite{LID} to describe the local adversarial regions based on the assumption that AEs are surrounded by several natural data subspaces. Alternatively, Lee \emph{et al.} used the MD to evaluate the dissimilarity between AEs and NEs \cite {OOD}. Furthermore, Xu \emph{et. al} \cite{FeatureSqueezing} devised a detection method via FS, such as reducing the color depth of images or  using smoothing. A similar strategy by using the non-linear dimension reducing technique was given by Crecchi \emph{et. al} \cite{DectTsne}.

\section{Sensitivity Inconsistency and Dual Classifier Design}

Fawzi \emph{et al.} pointed out the adversarial direction usually points to the highly-curved region of the decision boundary, and many examples share the same potential adversarial directions \cite{ClassificationRegion, AUP}. This results in a phenomenon that AEs are more likely concentrated in some common highly-curved regions. Imagine that we can fluctuate the highly-curved region of the decision boundary. Then AEs would be very sensitive to such fluctuations in the sense of producing different classification results. In contrast, NEs would be much less sensitive and generate quite consistent results under these fluctuations \cite{ClassificationRegion}. A simple yet illustrative example is given in Fig. \ref{fig:toy}, where the black line represents the decision boundary of a binary classifier and the red line shows the fluctuated (transformed) decision boundary. Here, the decision boundary transformation is conducted by flattening the highly-curved region, while keeping the remaining unchanged. In this case, under the transformed decision boundary, the classification results of AEs would change, while NEs exhibit quite consistent behaviour. This motivates us to exploit the distinct behaviour of AEs and NEs against the decision boundary transformation to tell them apart.

In the upcoming two subsections, we first formulate an optimization problem for finding an expected decision boundary transformation or equivalently the dual classifier. Then, we provide theoretical proofs and empirical justifications to demonstrate the effectiveness of the designed dual classifier.

\subsection{Design Dual Classifier in Transform Domain}

For the primal classifier  {\small{$\CF$}} trained on the dataset  {\small{$\BX$}}, let  {\small{$\CB_{i,j}$}} be the decision boundary between class $i$ and class $j$. Let also  {\small{$\tilde \CB_{i,j}$}} be a transformed version of   {\small{$\CB_{i,j}$}} corresponding to the dual classifier  {\small{$\CG$}}. We expect that  {\small{$\CB_{i,j}$}} is similar to  {\small{$\tilde \CB_{i,j}$}} except for those highly-curved parts. Apparently, the shape of  {\small{$\CB_{i,j}$}} highly correlates with the prediction confidence of  {\small{$\CF$}} on the training samples. Therefore, the problem of ensuring  {\small{$\tilde \CB_{i,j}$ and $\CB_{i,j}$}} to be similar except for highly-curved parts reduces to make  {\small{$\CF$}} and  {\small{$\CG$}} generate similar prediction confidences on NEs, but vastly different ones on AEs. Considering all the boundaries  {\small{$\CB_{i,j}$}}'s, we formulate the following optimization problem to search for  {\small{$\CG$}}:

\begin{equation}\label{optG}
\begin{aligned}
& \min_{\CG}  \max_{\Bx \in \BX} \left \| \CF(\Bx) - \CG(\Bx) \right\|^2_2,\\
& \text{s.t.} \left \| \CF(\hat \Bx) - \CG(\hat \Bx) \right\|^2_2 \geq \xi,  \quad \forall \hat \Bx \in \hat \BX
\end{aligned}
\end{equation}
where the cost function is designed to minimize the worst-case deviation of {\small{$\CG$}} from {\small{$\CF$}} on NEs, and the constraint guarantees their prediction confidences on AEs to be significant enough.

\begin{figure*}[!th]
	\centering
	\subfigure[]{\label{fig:P} \includegraphics[width=0.3\textwidth]{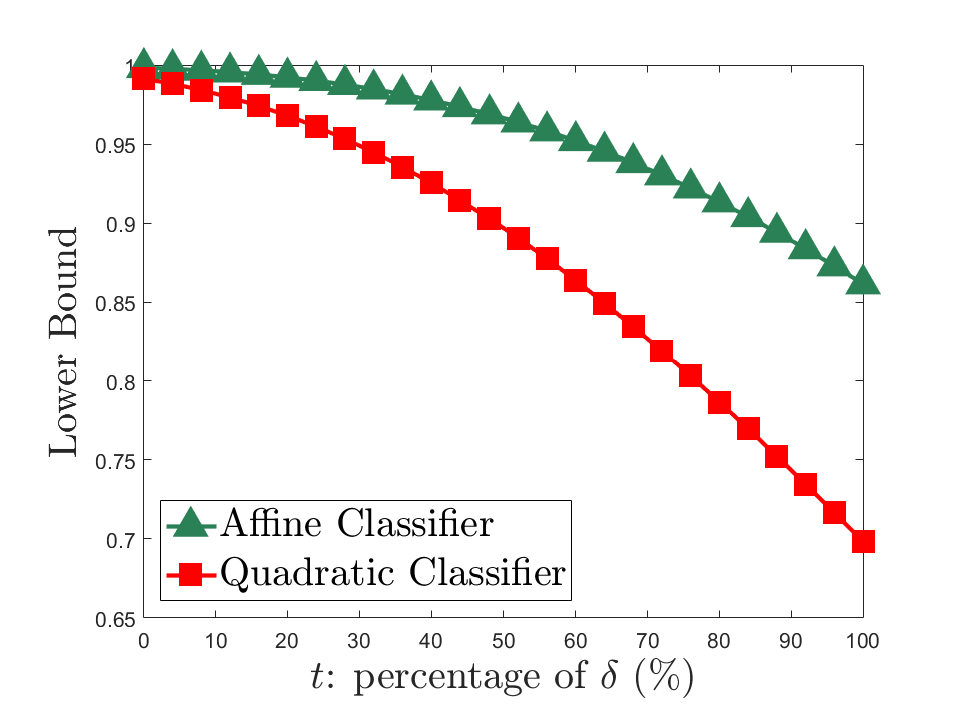} }
	\subfigure[]{\label{fig:VGGLoss} \includegraphics[width=0.3\textwidth]{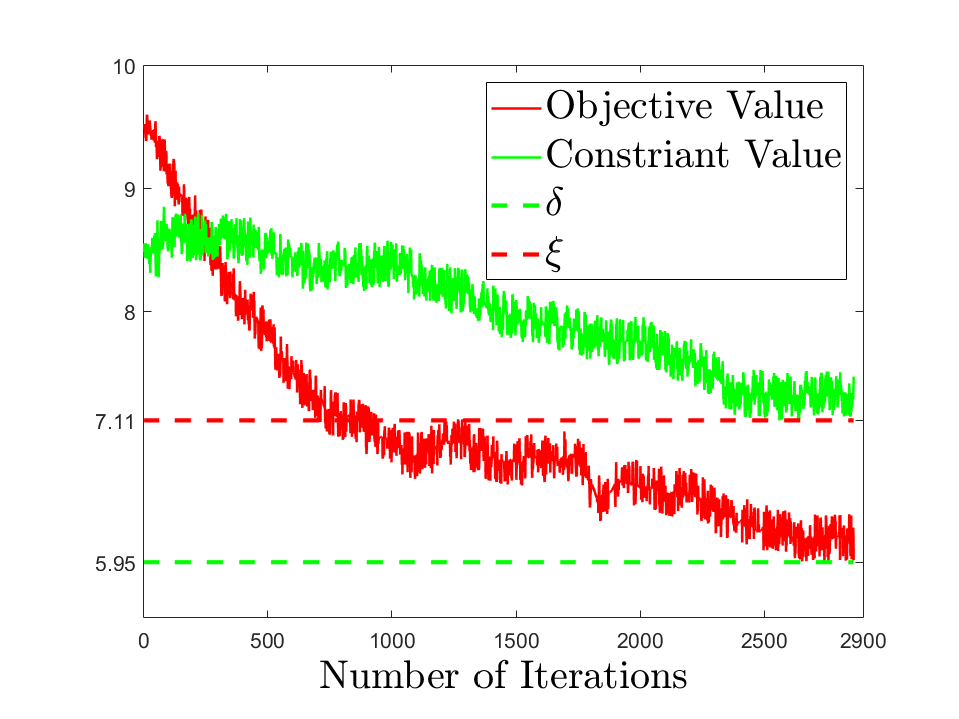}}
	\subfigure[]{\label{fig:vis}  \includegraphics[width=0.3\textwidth]{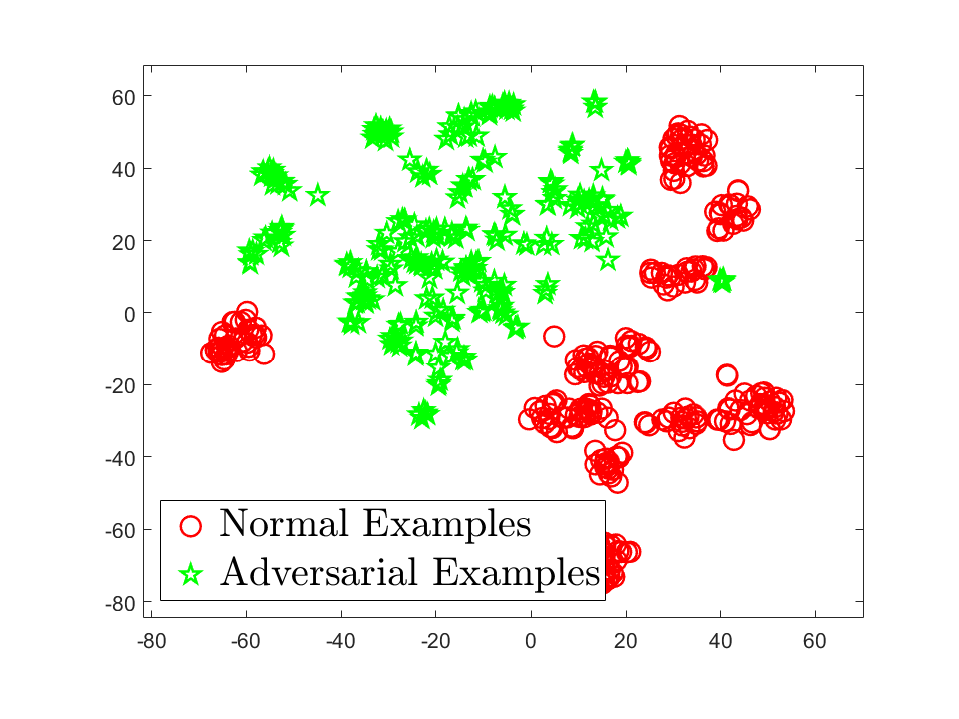}}
	\caption{Illustrations of the approximation precision of the constructed dual classifier {\small{$\CG$}}, and the separability of the feature of sensitivity inconsistency. (a) Lower bounds in (\ref{PAffine}) and (\ref{PQ}), (b) Objective and constraint values of (\ref{optG}) during the training process, and (c) the separable AEs and NEs.}
\end{figure*}

A straightforward way for solving this problem is to train the classifier {\small{$\CG$}} by integrating the constraint into the loss term. However, such naive solution easily causes overfitting \cite{overfit1,overfit2}. In this work, we resort to a transform-domain technique for getting an approximated solution of (\ref{optG}). An intuition why we search for  {\small{$\CG$}} in the transform domain is because DNN classifiers trained in the same domain tend to produce similar decision boundaries including the highly-curved parts  \cite{GeometricTransferability,fawzi2018nips,Goodfellow2017space}. A crucial problem now is how to appropriately choose the domain transform. Fawzi \emph{et. al} found that the decision boundary near NEs is flat along most directions, with merely few directions being highly-curved \cite{ClassificationRegion}. To maintain these flat parts, so as to preserve the results on NEs, we firstly could use linear transforms. Secondly, we should simultaneously flatten the highly-curved parts, which could be achieved by introducing distortions. The theory of manifold learning indicates that the distortion of the geometrical structure is caused by the dimensionality difference between the ambient and intrinsic spaces \cite{Distortion2, Mine}. Owing to these two reasons, we propose to design a linear, dimension-reducing transform, based on which we search for the desired  {\small{$\CG$}}. To this end, we define the transformation WAWT below. Let  {\small{$\Bx \in \BX$}}, and its WAWT transformed version  {\small{$\tilde{\Bx}$}} can be expressed as:

\begin{equation} \label{WAWT}
\tilde{\Bx} \triangleq  \mathbf{H} \Bx  = w_1 \Bx_{ll}+w_2 \Bx_{lh} +w_3 \Bx_{hl}  +w_4 \Bx_{hh},
\end{equation}
where {\small{$\Bx_{ll}$, $\Bx_{lh}$,$\Bx_{hl}$}}, and {\small{$\Bx_{hh}$}} represent the four sub-bands of a wavelet transform. Here, $w_i$'s are used to balance the importance of different sub-bands. Clearly, due to the lower dimension of each sub-band, {\small{$\BH$}} is a linear, dimension-reducing transform, as we expect. More details on WAWT can be found in the supplementary material.

Now, we can construct a classifier {\small{$\TCF(\TBx)$}} in the WAWT domain, and then the dual classifier can be naturally designed as {\small{$\CG(\Bx) = \TCF(\mathbf{H}\Bx)$}}. Training the dual classifier {\small{$\CG$}} is rather straightforward, by noticing that {\small{$ \CG$}} is the composition of {\small{$\mathbf{H}$}} with {\small{$\tilde \CF$}}. As shown in Fig. \ref{fig:framework} (the part enclosed by the red dashed line), {\small{$\CG$}} could be readily implemented by a DNN composed of the concatenation of {\small{$\tilde \CF$}} and WAWT layers with four trainable $w_i$'s. To guarantee that {\small{$\CG$}} and {\small{$\CF$}} have similar classification performance on NEs, the network structure of {\small{$\TCF$}} is exactly the same as {\small{$\CF$}}, except the input dimension. Note that the primal and the dual classifiers are trained separately.

Before diving into the design of the AE detector, let us first provide some theoretical analyses on the constructed dual classifier.

\subsection{Theoretical Justifications of Dual Classifier}

We aim to demonstrate: \textbf{if the predication confidence of the dual classifier {\small{$\CG$}} constructed based on the WAWT {\small{$\BH$}} is consistent with that of the primal classifier {\small{$\CF$}} on NEs, then {\small{$\CG$}} can approximate the optimal solution of the problem (\ref{optG}) with a given $\xi$}. We first restrict our theoretical analysis to affine and quadratic classifiers, as most of existing theoretical works assumed \cite{AtoR, ClassificationRegion}. We then offer empirical justifications for the general DNN classifiers. To make the analysis tractable, we assume that the dataset {\small{$\BX$}} is composed by {\small{$K$}} different classes of samples {\small{$\BX_{k}$ $(k = 1, ..., K)$}}, each of which is drawn from a $p$-dimensional Gaussian distribution {\small{$\mathcal{N}(\bmu_{k}, \BSG_{k})$}}.

For the affine classifier, we have the following theorem.

\begin{theorem}\label{Taffine}
	Let {\small{$\CF$}} be an affine classifier trained over {\small{$\BX$}} and {\small{$\CG$}} be the corresponding dual classifier based on the WAWT {\small{$\BH$}}. If
	
	\begin{small}
	\begin{equation}\label{T11}
	\max_{\Bx \in \BX} \left \| \CF(\Bx) - \CG(\Bx) \right\|^2_2 \leq \delta,
	\end{equation} 
	\end{small}then {\small{$\forall t>0$}} and {\small{$\hBx \in \hat\BX$}}, we have

	\begin{small}
	\begin{equation} \label{PAffine}
	\begin{aligned}
	&\mathbb{P}\left\{ \left \| \CF(\hBx) - \CG(\hBx) \right\|^2_2 \geq t + \delta \right\} \geq \\
	&\min_{k = 1,...,K} Q_{1/2}\left(|\BC_k(\frac{1}{2} \BI - \eta \BSG^{-1})\bmu_k|,\sqrt{(t+\delta)\slash {\BC_k \BSG_k \BC^T_k}}\right),
	\end{aligned}
	\end{equation}
	\end{small}where  {\small{$Q_{1/2}(\cdot)$}} is the Macrcum-Q function \cite{MQfun}, {\small{$\BSG$}} is the sample covariance matrix of {\small{ $\BX$, $\BC_k = \bmu^T_k(\BSG^{-1} - \BH^T(\BH\BSG\BH^T)^{-1}\BH)$}}, and $\eta$ is the perturbation magnitude.
\end{theorem}

Theorem 1 implies that, if the prediction confidence difference of the affine {\small{$\CF$}} and {\small{$\CG$}} is bounded by $\delta$, then the approximation precision of {\small{$\CG$}} to the optimal solution of problem (\ref{optG}) is determined by the probability in (\ref{PAffine}). To further illustrate the approximation precision, we randomly generate $10$-class mixture of Gaussian data {\small{$\BX$}}, and set $\eta=0.5$. Each class contains 100 examples of dimension 256.  Fig. \ref{fig:P} shows how the lower bound in (\ref{PAffine}) decays with respect to the increasing $t$. As can be observed, even when $t$ increases to $ 60 \% \delta$, this probability bound is still larger than 0.95, implying that the constructed $\CG$ is a precisely approximated solution of problem (\ref{optG}) with $\xi = 1.6 \delta$. In fact, Theorem \ref{Taffine} also reflects the necessity of constructing $\CG$ in the linear, dimension-reducing transform domain determined by {\small{$\BH$}}. If otherwise {\small{$\BH$}} is a dimension-preserving mapping, then {\small{$\BH$}} is an invertible matrix. In this case, {\small{$\BH^T(\BH\BSG\BH^T)^{-1}\BH = \BSG^{-1}$}}, making {\small{$\BC_k = \bm{0}, \forall k$}}. Hence, {\small{$\sqrt{(t+\delta)/\BC_k \BSG_k \BC^T_k} \rightarrow \infty$}}. Following the property of the Macrcum-Q function \cite{monotonyMQ}, the lower bound in (\ref{PAffine}) vanishes as the second term goes to infinity, regardless $t$ and $\sigma$. Besides, the Macrcum-Q function $Q_{1/2}(\cdot)$ in (\ref{PAffine}) increases as {\small{$|\BC_k(\frac{1}{2} \BI - \eta \BSG^{-1})\bmu_k|$}} does, which means that a large $\eta$ would result in a large lower bound in (\ref{PAffine}) and thus the designed dual classifier {\small{$\CG$}} is more precise.

For the case of quadratic classifier, we can similarly have Theorem 2.

\begin{theorem} \label{Tquadratic}
	Let {\small{$\CF$}} be a quadratic classifier and {\small{$\CG$}} be the corresponding dual classifier constructed by {\small{$\BH$}}. Both {\small{$\CF$}} and {\small{$\CG$}} are trained over {\small{$\BX$}}. If {\small{$\CF$}} and {\small{$\CG$}} satisfy (\ref{T11}), then $\forall t>0$ and {\small{$\hBx \in \hat\BX$}}, we have
	
	\begin{small}
	\begin{equation} \label{PQ}
	\begin{aligned}
	&\mathbb{P}\left\{ \left \| \CF(\hBx) - \CG(\hBx) \right\|^2_2 \geq t + \delta \right\} \geq \\
	&\min_{k = 1,...,K} \left\{ 1 - P(\sqrt{\delta + t}; \bm \lambda_{k}) + P(-\sqrt{\delta+t}; \bm \lambda_{k}) \right\} ,
	\end{aligned}
	\end{equation}
	\end{small}where {\small{$\bm \lambda_{k}$}}'s are the nonzero eigenvalues of the matrix
    
    \begin{small}
    \begin{equation} \label{QudaM}
	\hat \BSG_k^{\frac{1}{2}}(\BSG^{-1}_{k} - \BH^T(\BH\BSG_{k}\BH^T)^{-1}\BH)\hat \BSG_k^{\frac{1}{2}}
	\end{equation}
	\end{small}with {\small{$ \hat \BSG_k = (\BI + {\eta\BSG^{-1}_k}) \BSG_{k}(\BI + {\eta\BSG^{-1}_k})^T$, $P(\cdot, \bm \lambda_{k})$}} being the cumulative distribution function of the linear combination of Chi-square random variables with {\small{$\bm \lambda_{k}$}} as coefficients \cite{LinearCombinationChis}, and $\eta$ being the perturbation magnitude.
\end{theorem}

A similar curve on the lower bound in (\ref{PQ}) can be found in Fig.\ref{fig:P}, as in the affine case. From Theorem 2, we can also see that if {\small{$\BH$}} is invertible as a dimension preserving mapping, then the matrix (\ref{QudaM}) becomes a zero matrix with {\small{$\bm \lambda_{k} = \bm 0$, $\forall k$}}. Consequently, {\small{$1 - P(\sqrt{\delta + t}; \bm 0) + P(-\sqrt{\delta + t}; \bm 0) = 0$}} for any $\delta, t > 0$, resulting in a trivial bound in (\ref{PQ}). This again validates the usage of the linear, dimension-reducing {\small{$\BH$}}. Moreover, the lower bound increases as the increase of $\eta$. It becomes even clearer when each {\small{$\BX_{k}$}} is sampled from an independent multivariate Gaussian. In this case, the lower bound in (\ref{PQ}) reduces to {\small{$1-\mathcal{T}({3p}/{8},\sqrt{(t+\delta)/{4(1+\eta)^4}})$}}, where the regularized gamma function {\small{$\mathcal{T}(\cdot)$}} is decreasing as the increase of $\eta$ (see the Corollary 2 in the supplementary materials). The proofs of Theorems \ref{Taffine} and \ref{Tquadratic} are also given in the supplementary materials.
\begin{table*}[!tbp]
	\centering
	\begin{small}
		\begin{tabular}{c c c c c c c c c c c}
			\toprule
			Detector & Source     & DeepFool         &  FGSM            & BIM                      & C\&W              & Source & DeepFool      &  FGSM              & BIM      & C\&W  \\
			\midrule
			\hline
			SID          &\mrRC                           &$\bf{97.08}$    & 94.79                 & $\bf{99.38}$     & $\bf{96.56}$                 & \mrVC  & $\bf{98.32}$  & 84.92                 &$\bf{96.55}$  & $\bf{95.61}$\\
			MD        &                                       & 93.72                 & $\bf{99.89}$   & 92.73                    & 94.22                             &                & 93.84               & $\bf{99.76}$    & 89.13              & 78.84  \\
			LID           &                                       & 93.49                &96.54                 & 80.52                   &  81.21                               &                &90.02                &96.85                    & 80.04             & 73.68 \\
			FS            &                                        & 84.63                & 78.66                & 89.06                    & 85.51                               &              & 90.78               & 78.28                      & 79.94              & 76.34\\
			
			\cline{1-11}
			
			SID		     &\mrRS                          & $\bf{98.15 }$   & 93.64                   & $\bf{98.78}$   &$\bf{97.04}$                    &\mrVS    & $\bf{98.78}$ &94.41                             & $\bf{99.83}$  &$\bf{95.55}$ \\
			MD       &                                      &  95.43                 & $\bf{98.62}$     & 96.23                & 91.59                                &                & 87.31                & $\bf{99.73}$              &  90.79              &80.71 \\
			LID           &                                      & 92.43                 & 96.77                   & 90.77                  & 87.67                                &                & 91.51                & 98.32                            & 92.39               &83.07 \\
			FS            &                                      &95.72                  & 90.07                   & 93.66                  & 91.55                                 &                &93.99                 &82.97                           &  95.43             &87.52\\
			\hline
			\hline
		\end{tabular}
	\end{small}
	\caption{Comparison of AUC scores (\%) of detecting AEs under different settings. }
	\label{tab:Detect}
\end{table*}
\begin{figure*}[!t]
	\centering
	\subfigure[]{\label{fig:robust_FGSM} \includegraphics[width=0.3\textwidth]{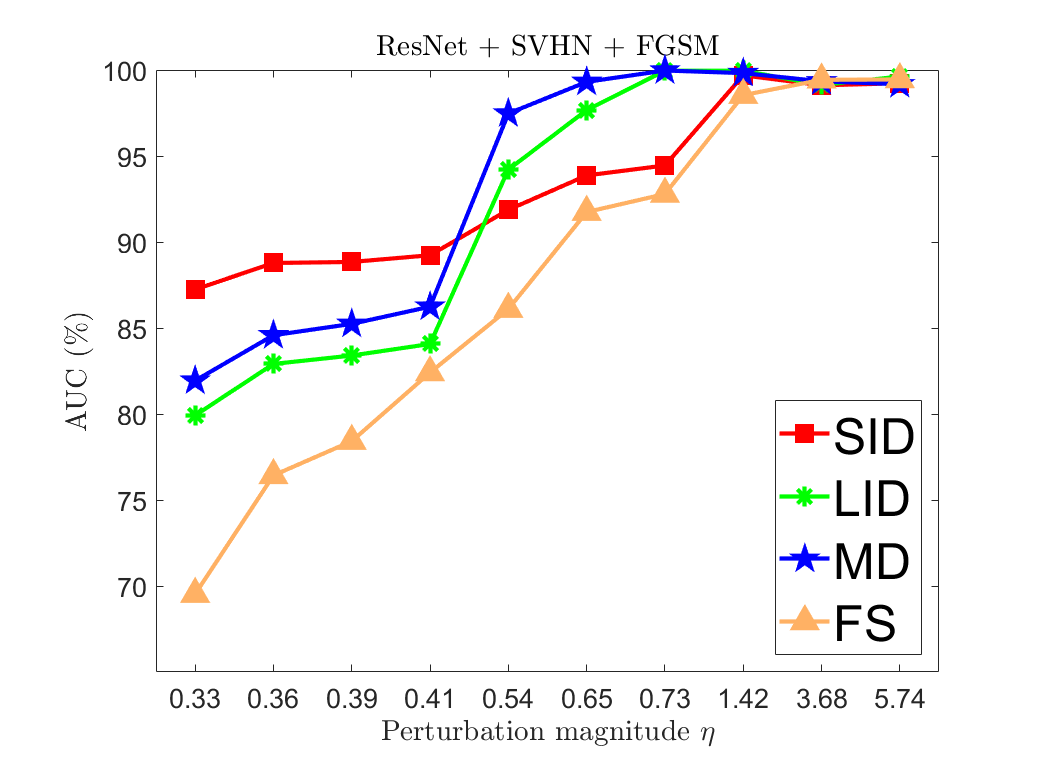} }
	\subfigure[]{\label{fig:robust_BIM} \includegraphics[width=0.3\textwidth]{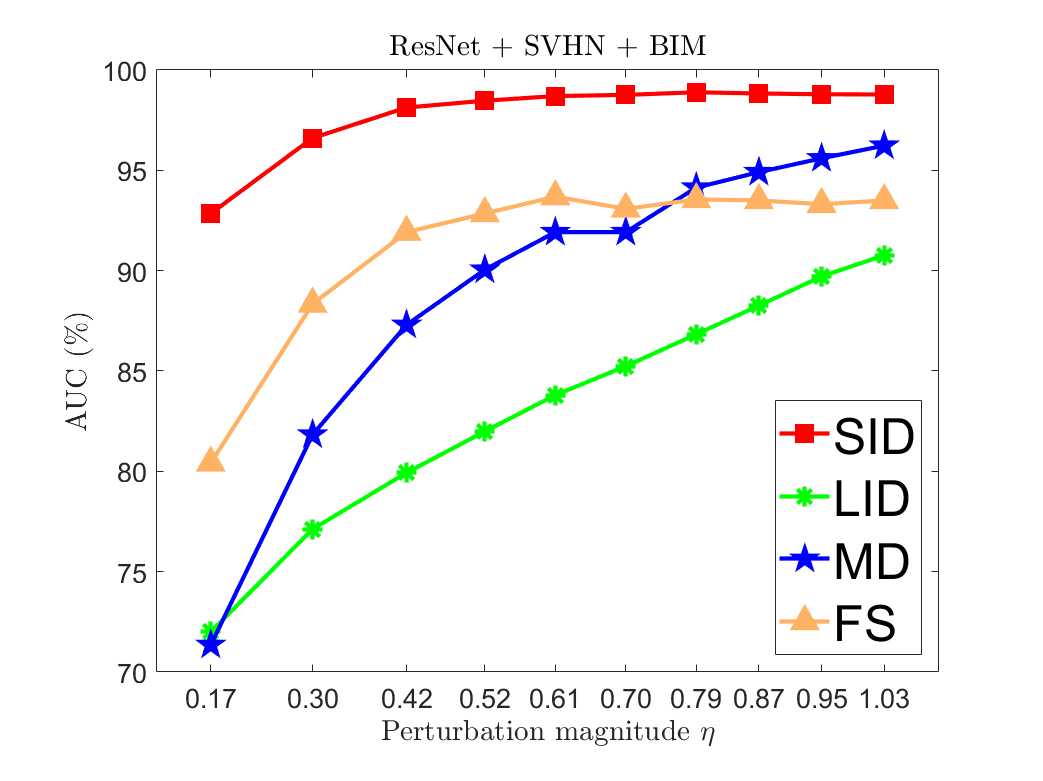}}
	\subfigure[]{\label{fig:robust_DF}  \includegraphics[width=0.3\textwidth]{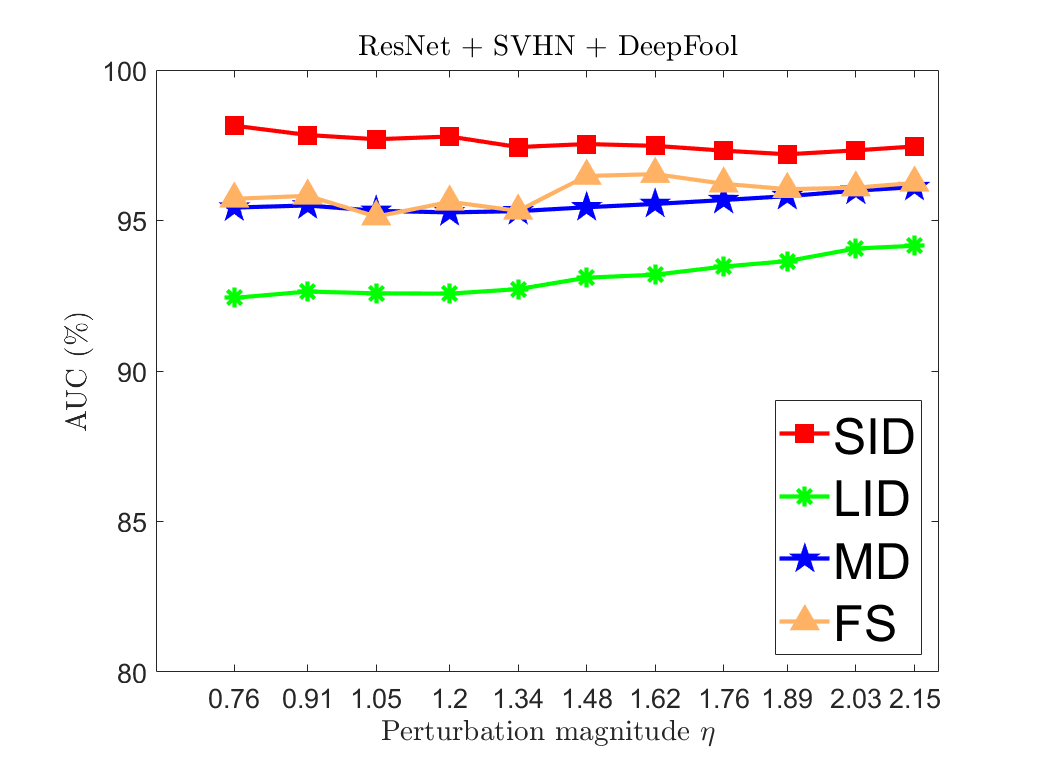}}
	\caption{Detection performance of SID, LID, MD, and FS w.r.t the perturbation magnitude $\eta$. Attacking (a) ResNet on SVHN with FGSM, (b) ResNet on SVHN with BIM, and (c) ResNet on SVHN with DeepFool. }
	\label{fig:robust}
\end{figure*}

In addition to these theoretical proofs, we give the empirical justifications on the effectiveness of the constructed {\small{$\CG$}}, for general DNN classifier with complex decision boundaries. As an example, we choose the VGG classifier on CIFAR10 as the primal classifier {\small{$\CF$}}, and then separately train the dual classifier {\small{$\CG$}}. During the training process of {\small{$\CG$}}, we calculate the values of the objective function in problem (\ref{optG}) on 500 test images, and the values of the constraint on their AEs produced by DeepFool \cite{DeepFool}. Fig. \ref{fig:VGGLoss} shows how the objective function and the constraint vary with the number of iterations of mini-batches. As can be observed, {\small{$\CG$}} tends to approximate the optimal solution of the problem (\ref{optG}) with $\xi = 7.11$, in which case the objective function is minimized to 5.95.

\begin{algorithm}[!t]
	\begin{small}
		\caption{Training procedure of SID}
		\KwIn{$\CF$: primal classifier; $\CG$: dual classifier; $\BX_c$: NEs correctly classified by $\CF$;
			$\mathcal{A}$: attack method; $E$: maximal training epoch.}
		\KwOut{$\mathcal {D}$: trained SID.}
		\While{$t<E$}{
			\For {$\mathbf{B}_{c}$ $\mathbf{in}$ $\BX_c$}{
				\mbox{$\mathbf{B}_{a} := $ attack $\mathbf{B}_c$ with $\mathcal{A}$;} \\ 
				\mbox{$\mathbf{B}_{n} := $ add random noise to $\mathbf{B}_c$;} 
				\mbox{$\triangleright$  $\mathbf{B}_{c}$: a minibatch of $\BX_c$.} \\ 
				\mbox{$\mathbf{D}_{1}:=$  $\left \{\Bx \in \{\mathbf{B}_{n}, \mathbf{B}_{c}\} |  k_{\CF}(\Bx) = k_{\CG}(\Bx) \right\}$;} \\
				\mbox{$\mathbf{D}_{2}:=$  $\left \{\Bx \in \{\mathbf{B}_{n}, \mathbf{B}_{c}\} |  k_{\CF}(\Bx) \neq k_{\CG}(\Bx) \right\}$;}
				\mbox{$\triangleright$  $k_{\CF}(\Bx), k_{\CG}(\Bx)$: predicted labels of $\CF$ and $\CG$.} \\
				\mbox{$\mathbf{D} := \{ \mathbf{D}_1, \mathbf{D}_2, \mathbf{B}_a\}$;} \\
				\mbox{$\mathcal{D}^{t}=$  $~ \min \limits_{\mathcal{D}^{t}} \frac{1}{|\mathbf{D}|} \sum_{\Bx \in \mathbf{D}} \mathcal{L} \left(\mathcal{D}^{t}(\CS(\Bx)),y \right)$;} 
				\mbox{$\triangleright$  $\mathcal{L}(\cdot)$: cross entropy.} 
				\mbox{$\triangleright$ $\CS(\Bx):$ calculate as (\ref{eq:SF}).} 
				\mbox{$\triangleright$ $\mathcal{D}^{t}(\CS(\Bx)) = \{d_0^t(\Bx), d_1^t(\Bx), d_2^t(\Bx)\}$.} 
				\mbox{$\triangleright$ $d_i^t({\Bx})$: the predicted confidences of $\mathcal{D}^{t}$.}\\
		}}
	\end{small}
\end{algorithm}

\section{Detecting Adversarial Examples from Sensitivity Inconsistency}

Upon the construction of the dual classifier $\CG$, we now collaboratively use it with the primal classifier $\CF$ to design a feature called sensitivity inconsistency. We also give the details on the proposed \emph{Sensitivity Inconsistency Detector} (SID) for discriminating NEs and AEs.

\subsection{The Feature of Sensitivity Inconsistency}
For a given unknown example $\Bx_0$, a natural measure to evaluate its sensitivity against the decision boundary transformation can be defined as:

\begin{equation}\label{eq:SF}
\CS(\Bx_0) = \Big \{f_i (\Bx_0) - g_i (\Bx_0) \Big \}^{K}_{i =1},
\end{equation}
where {\small{$f_i(\Bx_0)$}} and {\small{$g_i(\Bx_0)$}} are the associated prediction conferences under {\small{$\CF$}} and {\small{$\CG$}}, respectively. The vectorized {\small{$\CS(\Bx_0) $}} is then called the feature of sensitivity inconsistency of {\small{$\Bx_0$}}. The design process of {\small{$\CG$}} naturally endows the sensitivity inconsistency with the power to discriminate NEs and AEs. More precisely, {\small{$\forall \Bx \in \BX, \hat \Bx \in \hat \BX$}}, if {\small{$\CG$}} is the solution of problem (\ref{optG}), then there exists a gap between {\small{$\left \| \CF(\hBx) - \CG(\hBx) \right\|_2$}} and {\small{$\left \| \CF(\Bx) - \CG(\Bx) \right\|_2$}}. In other words, the inequality {\small{$\|\CS(\hat \Bx)\|^2_2 - \|\CS(\Bx)\|^2_2 \geq \epsilon $}} always holds for a certain $\epsilon$. In this case, there would exist a sphere such that {\small{$\CS(\Bx)$}} is located at the inside of it, while {\small{$\CS(\hat \Bx)$}} is distributed on the outside. Therefore, $\Bx$ and $\hat \Bx$ can be well separated.

To further show the separability of NEs and AEs based on the feature of sensitivity inconsistency, we give an example in Fig. \ref{fig:vis}, where the considered {\small{$\CF$}} and {\small{$\CG$}} are VGG trained on CIFAR10 and its dual version. We randomly select 300 correctly classified examples from the test set and generate their adversarial counterparts using DeepFool. Then, we use the algorithm T-SNE \cite{TSNE} to visualize the feature of sensitivity inconsistency of selected AEs and NEs. As can be observed, the feature vectors corresponding to them are largely separable. This example further motivates us to use the feature of sensitivity inconsistency for distinguishing AEs from NEs.

\subsection{The Design of SID}


We now give the details of SID, by virtue of the developed feature of sensitivity inconsistency. The schematic diagram of SID is depicted in Fig. \ref{fig:framework}, which consists of three parts: the pre-trained primal classifier {\small{$\CF$}}, the dual classifier {\small{$\CG$}}, and a detector trained on features of sensitivity inconsistency. For a given {\small{$\CF$}}, the dual classifier {\small{$\CG$}} is the composition of a WAWT layer and a DNN classifier with the same structure as {\small{$\CF$}}. The output prediction confidence of {\small{$\CF$}} and {\small{$\CG$}} is then passed into the detector which is composed of two fully connected layers. To improve the discrimination capability of SID, we design the detector as a three-class classifier; in addition to AEs (label 0) and NEs (label 1), those examples which are inconsistently classified by {\small{$\CF$}} and {\small{$\CG$}} are categorized into the third class (label 2).

The details of training the SID is given in Algorithm 1, which starts with the generation of adversarial and noisy counterparts to NEs. Similar to prior studies \cite{LID, OOD}, these NEs ({\small{$\BX_c$}}) are assumed to be correctly classified by {\small{$\CF$}}. For one mini-batch of NEs ({\small{$\BB_c$}}), the corresponding AE mini-batch ({\small{$\BB_a$}}) is generated using an attack strategy {\small{$\mathcal{A}$}} (line 3). The noisy mini-batch  ({\small{$\BB_n$}}) can be readily produced by adding random noises with the same noise magnitude as the adversarial perturbation added on {\small{$\BB_a$}} (line 4). Examples in {\small{$\BB_c$}} and {\small{$\BB_n$}} are further split into {\small{$\BD_1$}} which contains the examples consistently classified by {\small{$\CF$}} and {\small{$\CG$}}, and {\small{$\BD_2$}}, which corresponds to the inconsistently classified ones (line 5-6). We now have prepared a batch of examples {\small{$\BD$}} composed by three classes of examples, {\small{$\BB_a$, $\BD_1$, and $\BD_2$}} whose labels are $0$, $1$, and $2$ respectively (line 7). Upon having {\small{$\BD$}}, we can calculate the sensitivity inconsistency {\small{$\CS(\Bx)$}} of each {\small{$\Bx \in \BD$}} via (\ref{eq:SF}), and minimize the average cross entropy of SID on {\small{$\BD$}} (line 8). After exhausting all mini-batches in {\small{$\BX_c$}}, and repeating the above process for {\small{$E$}} times, we finally obtain the trained SID.

At the testing stage, classes 1 and 2 are merged. That is, a given example $\mathbf{x}_0$ is identified as an AE if the predicted label from SID is $0$; otherwise, it is treated as a NE.

\section{Experimental Results}

We evaluate the performance of the proposed SID on detecting AEs. We compare SID with the state-of-the-art schemes based on LID \cite{LID}, MD \cite{OOD}, and FS \cite{FeatureSqueezing}. In addition, we discuss the robustness of SID under a white box attack. We consider two network structures VGG19 with batch normalization \cite{VGG} and ResNet34 \cite{resnet} on two datasets CIFAR10 \cite{Cifar10} and SVHN \cite{SVHN}. The metric for evaluating the performance of different detectors is the widely-used AUC score. For fair comparison, all the parameters in LID, MD, and FS are fine tuned to achieve the best performance. In the WAWT, the wavelet transform $\bf{sym 17}$ is adopted \cite{pywt}.

\begin{table*}[!tbp]
	\centering
	\begin{small}
		\begin{tabular}{ c|c| c c c c}
			\toprule
			\hline
			\multirow{2}{*}{Model-Dataset} & \multirow{2}{*}{Source} &  \multicolumn{4}{c}{Target (SID/ MD / LID) }\\
			\cline{3-6}
			&                                 & DeepFool     &FGSM             &BIM                     & C\&W \\
			\hline
			\multirow{4}{*}{\tabincell{c}{ResNet\\ \& \\SVHN}}
			&DeepFool              & $\bf{98.15}$/95.32/87.26                     &$\bf{94.87}$/87.58/77.65       &$\bf{95.73}$/90.09/76.75           & $\bf{96.16}$/91.56/73.43\\
			
			&FGSM                     & 92.47 /$\bf{94.34}$/84.26            &$\bf{91.33}$/85.34/78.43        & $\bf{96.93}$/89.55/73.03          &$\bf{94.73}$/90.41/80.41\\
			
			&BIM                         &$\bf{95.62}$/94.65/90.87                      & $\bf{96.42}$/86.72/76.86       &$\bf{99.47}$/89.69/83.31          & $\bf{98.47}$/91.13/86.26\\
			
			&C\&W                      & $\bf{97.07}$/95.02/82.05                      & $\bf{96.06}$/87.54/78.57     &$\bf{98.06}$/89.67/74.52          & $\bf{97.87}$/91.52/77.68\\
			\hline
			\hline
			\multirow{4}{*}{\tabincell{c}{ResNet\\ \& \\CIFAR10}}
			&DeepFool             & $\bf{95.99}$/92.63/80.94 &$\bf{95.23}$/85.42/83.61  &$\bf{92.99}$/86.89/77.52     & $\bf{93.98}$/91.14/82.79\\
			
			&FGSM                     & $\bf{92.66}$/91.43/83.21 &$\bf{93.64}$/83.21/76.54 & $\bf{90.31}$/82.31/75.43     &$\bf{93.27}$/89.43/81.34\\
			
			&BIM                        &$\bf{94.26}$/90.99/83.21  & $\bf{95.27}$/84.35/79.31  &$\bf{99.68}$/87.01/79.39    & $\bf{99.31}$/92.65/83.04\\
			
			&C\&W                      & $\bf{96.02}$/90.91/82.51  & $\bf{96.37}$/84.32/79.13   &$\bf{95.54}$/87.25/78.53    & $\bf{97.58}$/93.78/87.53\\
			\hline
		\end{tabular}
	\end{small}
	\caption{Comparison of the generalizability (AUC scores).}
	\label{tab:generiablity}%
\end{table*}%
\begin{table}[!tbp]
	\centering
	\begin{small}
		\begin{tabular}{c c c c c}
			\hline
			\multirow{2}{*}{Source} & \multicolumn{2}{c }{RseNet-CIFAR10} & \multicolumn{2}{c}{RseNet-SVHN} \\
			\cline{2-5} & $A_O$    & $A_W$     & $A_O$     & $A_W$ \\
			\hline
			BIM-L & 99.81 & 98.31 & 98.51 & 98.13\\
			\hline
			BIM-S  & 97.17 & 96.31 & 95.65 & 95.13\\
			\hline
			FGSM-L & 96.13 & 93.47 & 98.37 & 96.53\\
			\hline
			FGSM-S & 87.93 & 87.74 & 83.88 & 82.35\\
			\hline
		\end{tabular}
	\end{small}
	\caption{Robustness of SID against the white box attack.}
	\label{tab:robustness}%
\end{table}%
\subsection{Detection Performance Comparison and Analysis}
We first show the performance comparison when AEs for training and testing are all generated by the same attack. In Table \ref{tab:Detect}, we compare the AUC scores of different detectors, where the attack methods DeepFool \cite{DeepFool}, FGSM \cite{FGSM}, BIM \cite{BIM}, and C\&W \cite{CW} are considered. The perturbation magnitudes $\eta$ of all the AEs can be found in the supplementary file. Compared with LID, MD, and FS, our proposed SID outperforms them in most cases, especially when the attack methods are more sophisticated, \eg, BIM, DeepFool, and C\&W.

To more thoroughly show the advantages of SID, we give the AUC scores of different detectors on AEs generated by FGSM, BIM and DeepFool, with respect to the varying $\eta$. As can be seen from Fig. \ref{fig:robust}, SID achieves better AUC performance than all the competing detectors on AEs generated by BIM and DeepFool, for all $\eta$'s. In the cases of AEs produced by FGSM, our SID is still the best when $\eta \leq 0.41$, and gradually becomes inferior to LID and MD when $\eta$ further increases. Arguably, when $\eta$ is large, FGSM could be deemed as unsuccessful, as the incurred distortions would be too large. In fact, when $\eta$ is significant enough (\eg, $\eta > 3.68$), the AUC performance of all detectors is almost perfect in the case of FGSM. This observation, in turn, reflects that the detection challenges mainly lie in the regime of small $\eta$, in which SID is the best detector in all the tested cases.

We also compare the generalizability of different detectors in Table \ref{tab:generiablity}. Note that here FS is not considered, as the training of FS merely relies on NEs. It can be found that the generalizability of SID is much improved in most cases, compared with MD and LID. Such desirable generalizability is attributed to the fact that AEs in the same curved regions would have similar sensitivity to the boundary transformation. For those sophisticatedly designed attacks such as DeepFool, BIM, and C\&W, the generated AEs can be regarded as good approximations to the optimal ones (in the sense of (\ref{optimal})). This implies that these AEs from different sources would be concentrated in the same curved regions as the underlying optimal ones, and hence, would exhibit similar sensitivity inconsistency. Hence, naturally, SID trained on one source is very likely to be able to detect AEs from other sources.

\subsection{Robustness of SID under white box attack}

We now evaluate the robustness of SID under the white box attack. In this scenario, an attacker knows everything including model parameters, training dataset, details on SID, etc. The purpose is to generate AEs to mislead the target classifier and simultaneously fool SID. A widely-adopted white box attack strategy is to  maximize the cost function {\small{$ \CL$}} of the classification model and loss $l$ of the detector at the same time \cite{OOD, LID}. Namely,

\begin{equation}
\begin{aligned} \label{WhiteBoxAttack}
& \max_{\Br} \CL(\hBx,y_c) + \alpha \cdot l(\hBx,y_d),  ~~\text{s.t.}~~\|\Br\|_2 < \eta,
\end{aligned}
\end{equation}
where $y_c$ is the ground-truth label of the NE $\Bx$, $y_d = 0$ indicates that $\hBx = \Bx + \Br$ is an AE, $\eta$ is the maximal allowable adversarial perturbation magnitude, and $\alpha$ is a tradeoff parameter.

To evaluate the robustness of SID, we randomly select 1000 NEs, which are correctly classified by ResNet, from CIFAR10 and SVHN respectively. We then generate their adversarial counterparts according to the white box attack strategy (\ref{WhiteBoxAttack}). The performance degradation of SID to the white box attack is examined in Table \ref{tab:robustness}. The first 2 rows give the results for SID trained on AEs produced by BIM with two different settings of $\eta$ (BIM-L and BIM-S for large and small $\eta$'s). The remaining two rows report the results for the cases of FGSM (FGSM-L and FGSM-S for large and small $\eta$'s). Detailed settings of $\eta$ can be found in the supplementary file. Here, {\small{$A_O$}} and {\small{$A_W$}} represent the AUC scores of detecting the original AEs and the ones generated by the white box attack. For instance, the SID trained on BIM-L source achieves $99.81\%$ AUC score on detecting original AEs targeted on ResNet-CIFAR10, while still obtaining $98.31\%$ on detecting AEs produced by the white box attack. As the AUC scores for detecting original AEs and the ones from (\ref{WhiteBoxAttack}) are close, we claim that SID shows promising robustness against the white box attack.

\section{Conclusions}

In this paper, we have proposed a simple yet effective method for detecting AEs, via the sensitivity inconsistency between NEs and AEs to the decision boundary fluctuations. Experimental results have been provided to show the superiority of the proposed SID detector, compared with the state-of-the-art algorithms based on LID, MD, and FS.

\section{ Acknowledgments}
This work was supported by Macau Science and Technology Development Fund under SKL-IOTSC-2018-2020, 077/2018/A2, 0015/2019/AKP, and 0060/2019/A1, by Research Committee at University of Macau under MYRG2018-00029-FST and MYRG2019-00023-FST, by Natural Science Foundation
of China under 61971476 and 62001304, and by Guangdong Basic and Applied Basic Research Foundation under 2019A1515110410.

\bibliography{myrefs}

\end{document}